# IMPLEMENTATION OF A VISION SYSTEM FOR A LANDMINE DETECTING ROBOT USING ARTIFICIAL NEURAL NETWORK

Roger Achkar and Michel Owayjan

Department of Computer and Communications Engineering, American University of Science & Technology, Beirut, Lebanon
{rachkar, mowayjan}@aust.edu.lb

***ABSTRACT***

*Landmines, specifically anti-tank mines, cluster bombs, and unexploded ordnance form a serious problem in many countries. Several landmine sweeping techniques are used for minesweeping. This paper presents the design and the implementation of the vision system of an autonomous robot for landmines localization. The proposed work develops state-of-the-art techniques in digital image processing for pre-processing captured images of the contaminated area. After enhancement, Artificial Neural Network (ANN) is used in order to identify, recognize and classify the landmines' make and model. The Back-Propagation algorithm is used for training the network. The proposed work proved to be able to identify and classify different types of landmines under various conditions (rotated landmine, partially covered landmine) with a success rate of up to 90%.*

***KEYWORDS***

*Artificial neural network applications; Back propagation algorithm learning; Image Segmentation; Landmines; Pattern Clustering*

## 1. INTRODUCTION

Landmines were mainly designed as area-denial weapons, and are used to create tactical barriers in order to prevent direct attack or to deny access by military and civilians from a defined area. Landmines are perfect soldiers that never eat, sleep, miss, fall ill or disobey. Moreover the landmine perfectly completes its job for much less U.S. dollars than the human soldier; in addition landmines are long-term killers, active long after a war has ended [1].

There are two major types of landmines, Anti-Personnel (AP) mines and Anti-Tank (AT) mines. AP mines are usually placed under earth, close to the surface; while AT mines are usually placed on the surface of the earth.

Accordingly to the UNICEF, there are an estimated 110 million active landmines buried in over 64 countries around the world. Around 2,000 persons are involved in monthly landmine accidents, 800 (40%) of whom are innocent civilians; that is, an average of a victim every 20 minutes dies [2]. According to the UN, even though about 100,000 mines are removed every year; two million more replace them [3], [4].

As for Lebanon, the Lebanese Army has estimated that there are about 406,000 landmines laid throughout the country, with South Lebanon being the most heavily contaminated area [5].

As such, in 2001, 85 landmine/UXO casualties, 14 killed and 71 injured, were reported in Lebanon. In 2002, this number showed a significant decrease: 48 casualties were reported among whom 16 persons died; 12 were killed during the demining process.

Since 1975, the devices have killed or maimed more than one million people worldwide; they kill 800 victims each month and disable more than 1,000 others [6].

## 2. MOTIVATION AND REQUIREMENTS

At the beginning of the 20th century, nearly 80% of landmine victims were military personnel. Today, 90% of landmine victims are civilians, most of whom are children who are more likely to be killed by mines than adults.

These mines, not only inflict physical and psychological damages on civilians, injure or kill considerable number of people a month, but also disturb the economic development of nations where buried mines abound, and prevent these countries from achieving socioeconomic stabilization. Landmines disrupt social services, threaten food security by preventing thousands of hectares of productive land from being farmed, and hinder the return and resettlement of refugees and displaced persons. A report from the United Nations High Commissioner for Refugees (UNHCR) published in 1997 stated that 13.2 million refugees, 4.9 million internally displaced people and 3.3 million returnees were at risk from landmines [7]. In addition, landmines cause environmental damage in the forms of soil degradation, deforestation, and the pollution of water resources with heavy metals.

Hence, the need to detect and remove landmines has become a necessity to preserve human life, as well as to the development of nations and their socioeconomic stabilization security.

## 3. STATE-OF-THE-ART SOLUTIONS

Minesweeping and removing landmines carry certain risks and can be slow and costly. Statistics by the Canadian Red Cross show that a single landmine can cost up to $30; whereas, removing a mine may cost up to $1,000 [8].

There are three kinds of minesweeping strategies ranging from a manual based minesweeping, a mechanical equipment based minesweeping, to an advanced robot based minesweeping.

The manual based technique relies on trained deminers sweeping the ground using metal detectors. Its speed is approximately 25 $m^2/h$. Metal detectors yield about one thousand false positives for every mine; this result is mainly due to the small size of the mines and the fact that many are made almost entirely of non-metallic materials. This technique is typically expensive, slow and dangerous [9], [10].

Mechanical based demining uses machinery to roll over the landmines and destroy them while they are still in the ground. This technique is known as the fastest demining technique. However, the machines employed in this approach are expensive to operate and can only be used when the terrain is suitable. Additionally, in most situations, this technique is not 100% reliable; thus, the need to employ another technique to check the minefield's clearance [9], [11].

Utilization of independent robots in order to detect landmines is still a process under development and is expected to be implemented in future's demining process. Employing an independent robot in the process of detecting mines will ensure the safety of local residents and those who are engaged in the minesweeping work and the demining process.

A landmine detecting robot sweeps the ground to detect the existence of a mine. The robot, through signal processing, decides whether a mine exists or not. The robot uses a sensor to capture real-time images of the area of interest. The captured images are then fed to a computing unit to be digitally processed before being analyzed. Digital Image processing techniques include noise filtering and image enhancement as well as segmentation and extraction of objects' patterns. Finally, these patterns are used to identify whether the detected object is a mine or not.
A robot may be an effective and efficient means of detecting mines; however, it may not be able to operate in all minefields since most minefields are found in unstructured terrain in uneasy to access areas.

Each solution has advantages and disadvantages. Table 1 provides a comparison between the different minesweeping solutions. The criteria of comparison are: human safety, accessibility to various locations, detection speed, and informative in terms of providing information regarding the detected mines.

According to the criteria mentioned above, the autonomous robot with machine vision is the best to fit the specifications set for the purpose of this research.

Table 1. Potential solutions comparison.

| **Minesweeping Technique** | **Human Safety** | **Accessibility** | **Detection Speed** | **Informative** |
|---|---|---|---|---|
| Manual Based Minesweeping | X | ✓ | x | x |
| Mechanical Based Minesweeping | ✓ | x | ✓ | x |
| Autonomous Robot | ✓ | x | ✓ | ✓ |

## 4. TECHNIQUES AND DESIGNS

This section gives a general overview of the operational process of the vision system of the autonomous landmine detecting robot.

The robot first captures an image in real-time. The image is then fed into a processing unit where it, or smaller segments of it, is (are) analyzed after being processed for noise reduction and enhancement. At the end, the imaged object(s), if any, is (are) recognized and classified. This general process is represented in the schematic drawing of Figure 1.

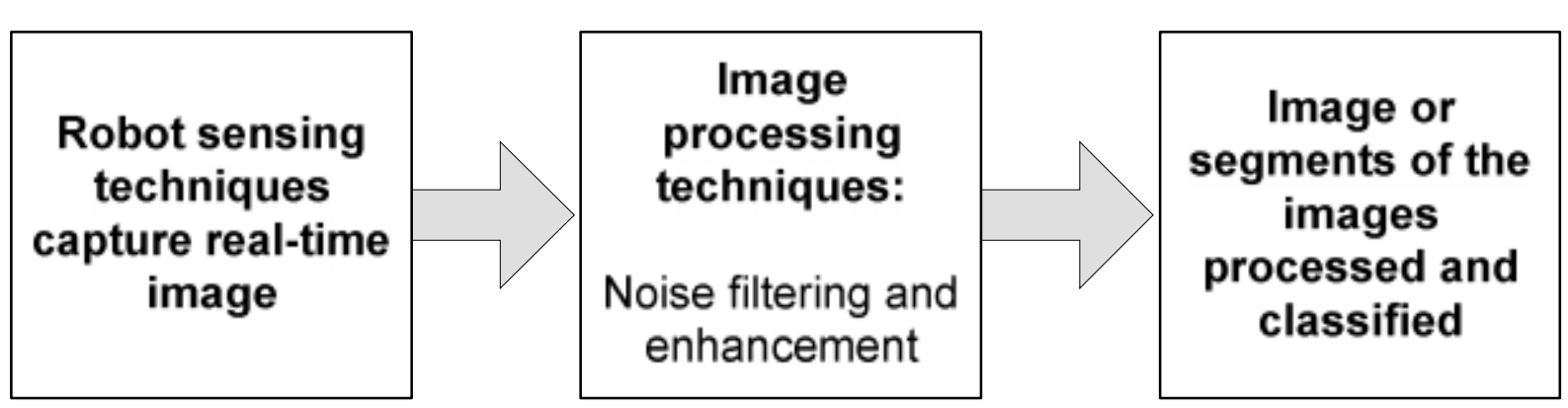

Figure 1. Schematic presentation of the general process

### 4.1. Robot's Detecting Techniques

The robot may employ several sensors in order to capture real-time images. The most common sensors are the Ground Penetration Radar (GPR) sensors, infrared (IR) sensors, and ultrasound (US) sensors, or cameras [12].

*Ground Penetrating Radar*: GPR may be used to detect landmines laid on the surface or buried underground. It works best in dry surfaces; however, it cannot be employed in high-conductivity areas such as soils that are salt contaminated or clay soils. Furthermore, the GPR has high energy consumption [13].

*Infra Red* (IR): The IR is also referred to as thermal radiation since the reflected radiation is detected in the form of heat. IR sensors are capable of detecting surface-laid mines as well as buried mines. The detection may be influenced by the type of soil especially since the water content in the soil may produce faulty detection. However, IR sensing does not require as much energy consumption as that of GPR [13].

*Ultrasound* (US): The US emits ultrasound signals and collects reflected signals from the surroundings. The US wave propagates well in humid or underwater conditions [13].
*Camera*: A camera may only be used in order to detect mines that are laid on the surface, whether in humid, dry, or any other conditions. Additionally, the image produced contains clearer details which may lead to a more accurate detection.

As shown in Table 2, both the camera and the US sensing techniques meet five out of the six set criteria. However, since the objective of this research is to detect AT mines, ground penetration is not a necessity, but the quality of the captured images is. Accordingly, this research adopts the camera as the sensing technique.

Table 2. Sensing techniques comparison.

| Sensing Technique | Cost | Energy Consumption | Quality of Images | Usability under Various Weather Conditions | Ground Penetration | Object Material Independency |
|---|---|---|---|---|---|---|
| Camera | ✓ | ✓ | ✓ | ✓ | x | ✓ |
| Infra Red | ✓ | ✓ | x | x | ✓ | x |
| Ultrasound | ✓ | ✓ | x | ✓ | ✓ | ✓ |
| Ground Penetrating Radar | x | x | x | x | ✓ | ✓ |

### 4.2. Image Processing Techniques

Captured information may be represented in the form of two dimensional data image where contrast in pixel intensity provides a clue for potential existence and location of mines. Digital image processing techniques include noise filtering and image enhancement as well as segmentation of objects.

#### 4.2.1. Noise Filtering

Any sensor is affected by a certain degree of noise that is unavoidable in the output of most sensors during digital image acquisition stage. In imaging, the noise may affect one pixel or several, and alter it (their) original value by an undefined amount [14]. Hence, the need to remove or diminish its effects.

There are several special filters that may be applied to remove noise and enhance the image; specifically, the Mean Filter, the Median Filter and the Gaussian Filter.
The Mean Filter, or averaging filter, is a low pass filter that smoothes local variation in an image. It simply replaces each pixel value in an image with the mean value of that of its neighbors, including itself. This process has the effect of eliminating pixel values which are unrepresentative of their surroundings.

The Median Filter is known as an order-statistics filter. It replaces the value of a pixel by the median of the levels in the neighborhood of that pixel; the original value of the pixel is included in the computation of the median.

The Gaussian smoothing Filter is used to remove details and noise; in this sense it is similar to the Mean Filter [15].

#### 4.2.2. Segmentation and Classification

The objective of segmentation is to divide the original image into sub-regions or smaller segments. These sub-regions correspond to individual surfaces, objects or natural parts of objects. Each segmented image is then fed into the computing unit to be analyzed and classified. Segmentation is a very important process as the segmentation accuracy determines the eventual success or failure of computerized analysis procedures.

Sapkal et al. [2] studied the K-Means algorithm and the Back-Propagation in detail and implemented them on various database for the segmentation purpose. It was found that K-Means algorithm gives very high accuracy, but it is useful for single database at a time; whereas neural network is useful for multiple databases, once it is trained. For the purpose of this study, the K-Means clustering is adopted as the segmentation algorithm in order to pre-analyze the image and prepare it for final investigation.

K-Means clustering is an unsupervised learning algorithm that targets to classify a given data set through a certain number of clusters (k). The main idea is to define k centers (centroids), one for each cluster. Pixels in the processed image are then classified and assigned to the nearest cluster. Each cluster represents a sub-region of the image. Assignment of the image pixels to clusters may be based on the nearest image pixel to the centroids in relation to color, texture, location, or a combination of several parameters.

### 4.3. Recognition Techniques

Object recognition is finding, identifying, and labeling a given object in an image. Recognition of objects in images should be accomplished despite the fact that the image of the objects may vary somewhat in different sightings, sizes/scale or even when the objects are translated or rotated.

#### 4.3.1. Matching using minimum distance classifier

Recognition based on matching assigns each unknown pattern to the class to which it is closest in terms of a predefined metric. The minimum distance classifier computes the Euclidean distance between the unknown and all the prototype patterns; the prototype pattern resulting in the smallest distance is then flagged as the recognized object.

Matching using the minimum distance classifier technique requires a lot of processing at run-time.

#### 4.3.2. Recognition by Neural Network

A neural network is a massively parallel distributed processor, made up of simple processing units, which has a natural tendency for storing experiential knowledge and making it available for use. A neural network is a computational model, inspired by the natural neurons, that processes information in a similar way to that of the human brain [16]. The artificial neuron resembles the natural neuron of that of the brain in two aspects: acquires knowledge through learning, and knowledge is stored within inter-neuron connection strengths known as synaptic weights.

A simple artificial neuron is a device with many inputs and one output. The neuron has two modes of operation: the training mode and the using mode. In the training mode, the neuron is trained to be activated (or not), for particular input patterns. In the using mode, when a taught input pattern is detected at the input, its associated output becomes the current output. Accordingly, an artificial Neural Network is made up of interconnecting artificial neurons.
The most common neural network model is the multilayer feedforward network which creates a model that correctly maps the input to the output using historical data. The model can then be used to produce the output when the desired output is unknown.

The multilayer feedforward neural network is a network where the neurons are organized in the form of layers. The network consists of an input layer of source nodes that projects onto another successive layer called the hidden layer. The hidden layers eventually project onto an output layer of neurons. The function of the hidden neurons is to intervene between the external input and the network output in some useful manner. With the use of one or more hidden layer, the network is enabled to extract higher-order statistics. This ability is particularly valuable when the size of the input layer is large.

The source nodes in the input layer of the network supply respective elements of the input vector in the second layer which is the first hidden layer. The output signals of the second layer are used as inputs to the third layer and so on for the rest of the network.

Back-Propagation is a common method of teaching the multilayer feedforward neural network how to perform a given task.

This learning technique is a supervised learning one, and is an implementation of the Delta rule. It requires a teacher that knows, or can calculate, the desired output for any given input [17].
This algorithm trains a given feed-forward multilayer neural network for a given set of input patterns with known classifications. When each entry of the sample set is presented to the network, the network examines its output response with respect to the sample input pattern. The output response is then compared to the known and desired output, and the error value is calculated. Based on the error, the connection weights are adjusted.

## 5. TECHNIQUES AND DESIGNS

A landmine detecting robot sweeps the ground and through signal processing, decides whether a mine exists or not. The robot first uses a sensor, such as a camera, to capture real-time images of the scanned area. The captured images are then fed to a computing unit to be digitally processed before being analyzed. Digital Image processing techniques include noise filtering and image enhancement as well as segmentation and extraction of objects' patterns. Finally, the extracted patterns are used to identify whether the detected object is a mine or not. Figure 2 provides a general overview of the implemented process.

The system uses the Artificial Neural Network in the recognition and classification process. Neural Network has the ability to learn from experience; however, it has to be taught. Accordingly, the system is trained to recognize and classify specific objects (landmines) [18].

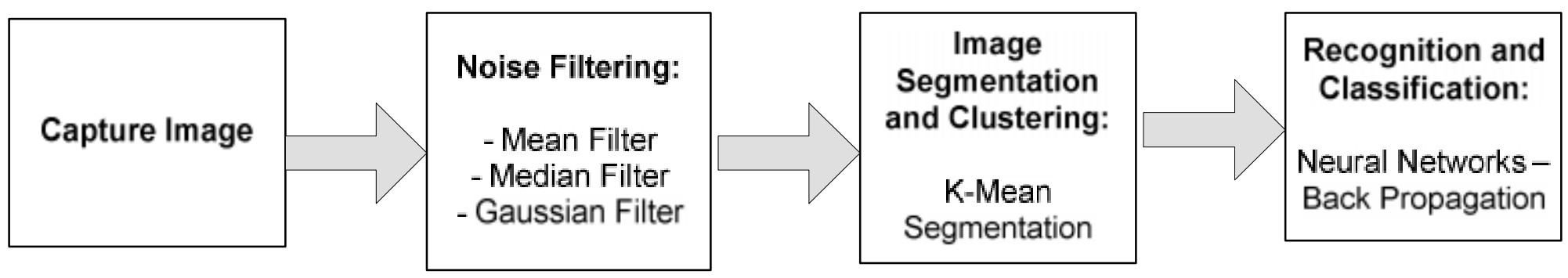

Figure 2. Schematic recognition and classification process

In order to train the system, a training data set needs to be prepared to be used as the database of the objects that the network is to identify. In order to establish the training data set, images containing the objects to be later recognized are acquired; these images are then processed, filtered and segmented in order to extract objects' patterns from the images. The extracted patterns, which are in the form of smaller images, are scaled into pre-defined size images. This process is described in Figure 3.

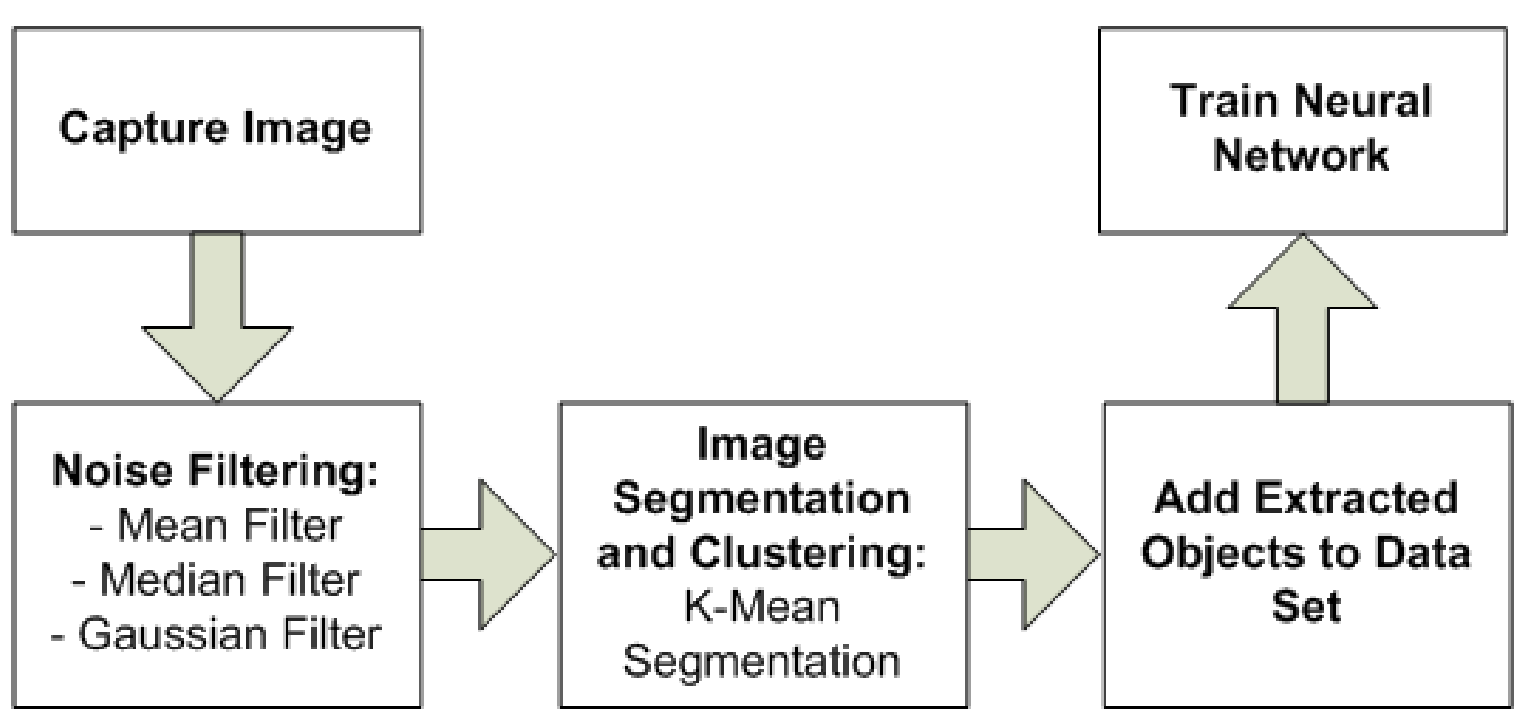

Figure 3. Training process within the Neural Network

### 5.1. Capture Image

The system uses the camera as a sensing technique to capture real-time images. The captured images are represented in the RGB color space.

## 5.2. Noise Filtering

As stated earlier, any sensor is affected by a certain degree of noise; hence, images need to be filtered.

To apply a filter on RGB images, the image is split into three grayscale images. The first created image consists of only the red components of the original RGB image; whereas the second and the third images consist of the green and blue components respectively. The filter mask is then applied separately to the temporarily created images. Finally, the tweaked grayscale images are concatenated into a new filtered RGB image.

### 5.2.1. Mean filter

In the mean filtering a 3×3 square mask, as shown in Figure 4, is often applied to every pixel of the image as depicted in (1); a 5×5 squares mask may be used for more severe smoothing. The process is simply moving the filter mask from point to point in an image. To get the resulting pixel at (x, y), the mask should be centered on the image at point (x, y) [15], [19].

$$
\begin{aligned}
d = {} & w_1 f(x-1, y-1) + w_2 f(x, y-1) + w_3 f(x+1, y-1) \qquad (1)\\
& + w_4 f(x-1, y) + w_5 f(x, y) + w_6 f(x+1, y)\\
& + w_7 f(x-1, y+1) + w_8 f(x, y+1) + w_9 f(x+1, y+1)
\end{aligned}
$$

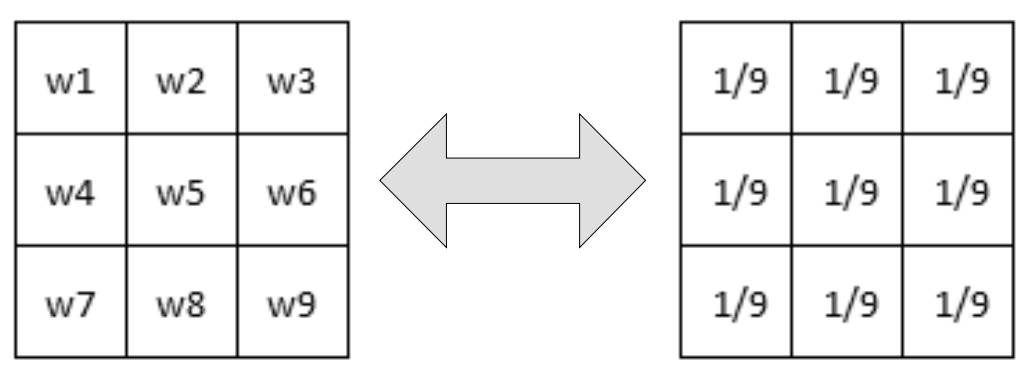

Figure 4. Mean 3x3 mask

### 5.2.2. Median filter

Through this filter the neighborhood pixels are processed by applying a mask, as shown in Figure 5, to the image. The implemented process is the same as that used with the Mean Filter. The median is calculated by first sorting all the pixel values from the surrounding neighborhood into numerical order, and then replacing the pixel being considered with the middle pixel value [15], [19].

| | | |
|---|---|---|
| 1 | 1 | 1 |
| 1 | 1 | 1 |
| 1 | 1 | 1 |

Figure 5. Median 3x3 mask

### 5.2.3. Gaussian filter

This filter is similar to the Mean Filter, but it uses a different kernel, as shown in Figure 6 [15], [19].

$$(1/16)\begin{bmatrix} 1 & 2 & 1 \\ 2 & 4 & 2 \\ 1 & 2 & 1 \end{bmatrix}$$

Figure 6. Gaussian 3x3 mask

## 5.3. Image Segmentation and Clustering

As stated, the K-Means clustering algorithm is used in this study to extract the different objects' patterns located in an image by categorizing a given data set through a certain number of clusters (k). The K-Means clustering is an iteration based algorithm; initial guesses are required to get the algorithm going, then recurring calculations are carried out until convergence.

As a next step, each point belonging to a given data set is taken and associated to the nearest centroid; the nearest centroid is determined by calculating the Euclidean distances between the points and the centers of each cluster, based on color, location, or a combination of several parameters. This study implements a combination of the color and the location of the pixel [20] to determine the Euclidean distance between the point and the center of the cluster as shown in (2).

$$d = \sqrt{\begin{array}{l}(R_{point} - R_{cent})^2 + (G_{point} - G_{cent})^2 + (B_{point} - B_{cent})^2 \\ + (X_{point} - X_{cent})^2 + (Y_{point} - Y_{cent})^2\end{array}} \quad (2)$$

Accordingly, the pixel under investigation is associated with the cluster with the smallest calculated distance. This process is computed for every pixel in the image. At this point, every pixel in the image is correlated to a cluster. Then new cluster centers are calculated based on the average of the data sets within each of the clusters. Once the new cluster centers are calculated, each pixel is recomputed in order to determine to which cluster it belongs [21].

Through the initial assignment of k and the centroids may be computed randomly or pre-assigned by the user; however, an automated process, the Simple Cluster-Seeking (SCS), is used in order to determine k. The SCS method is initialized by assigning the first instance or pixel of the image to a first class; this point will be referred to as a seed. The Euclidian distance between this first pixel and the next point is calculated. If the calculated distance is greater than a pre-defined threshold, then a new point is selected as the second seed; otherwise, the next pixel in the image is taken, and the process is repeated. The threshold, used to determine the seeds, is manually defined.

Once the second seed is chosen, the next pixel in the image is taken and the Euclidian distance between it and the two previous selected seeds is calculated. If both these distances are greater than the threshold, then the point to be the third seed is selected. The entire process is repeated until all pixels are processed. The number of clusters identified in the image would be the number of chosen seeds [22].

## 5.4. Training Data Set

Once an image is segmented and the different objects are determined, each identified object is separately considered as an image and added to the data set that is used to train the neural network.

The neural network requires the input vector to have a unique dimension, which is the total number of source nodes. Therefore, the newly created images need to be scaled to a pre-specified dimension (width x height). For this purpose, this study adopts a constant image dimension of (64 x 64) for a total of 4,096 pixels per object.

It is to be noted that when extracting the objects' patterns into new images of the same size of the original image and then transforming the images by scaling each to a 64 x 64 image, lots of information was lost due to the fact that the extracted object's patterns are relatively smaller than the size of the original image. Therefore, a blank needs to be cropped from the newly created image prior to the transformation by scaling.

Cropping an image consists of entirely removing the row or column of pixels with no levels, or any specified level. Detecting the blanked pixels and cropping the image will determine the region of interest in that image; specifically, the location of the identified object. In order to crop the image, a rectangle specifying the region of interest needs to be drawn. For this purpose, it is sufficient to determine the coordinates of the rectangle's corners as shown in Figure 7.

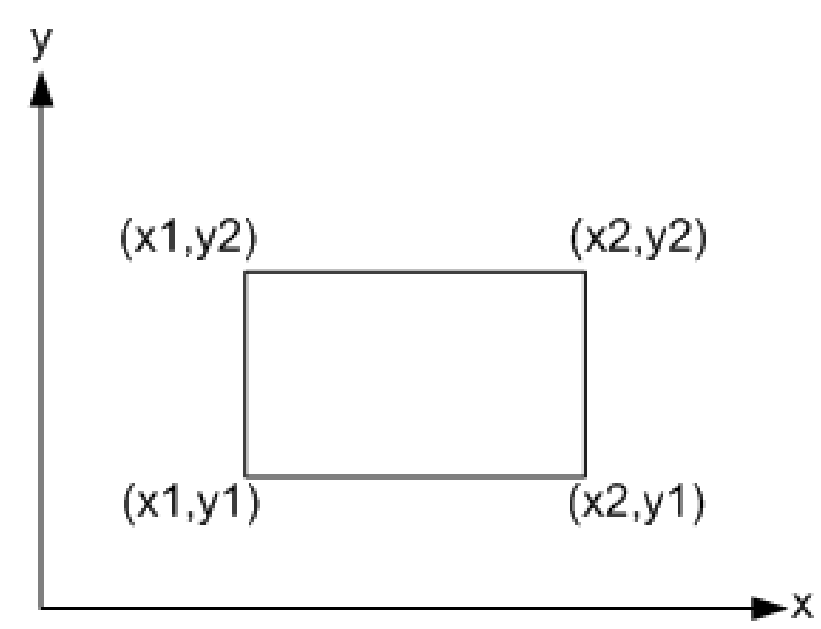

Figure 7. Region of interest rectangle

The process to perform the cropping is detailed herein:

- Starting from the top of the image to its bottom, the levels of each pixel in the row under investigation are added up. If the total is less than the specified level, y2 is set to be equal to the row's number under investigation. This is repeated for the next row. If the total of the pixel levels in the row is higher than the specified level, the next point is taken.
- Starting from left of the image to its right, the levels of each pixel at the column under investigation are added up. If the total is less than the specified level, x1 is set to be equal to the column's number under investigation. This is repeated for the next column. If the total of the pixel levels in the column is higher than the specified level, the next point is taken.
- Starting from bottom of the image to its top, the levels of each pixel at the row under investigation are added up. If the total is less than the specified level, y1 is set to be equal to the row's number under investigation. This is repeated for the next row. If the total of the pixel levels in the row is higher than the specified level, the next point is taken.
- Starting from right of the image to its left, the levels of each pixel at the column under investigation are added up. If the total is less than the specified level, x2 is set to be equal to the column's number under investigation. This is repeated for the next column. If the total of the pixel levels in the column is higher than the specified level, the calculations are not continued.

At this point, the coordinates of the rectangle's corners could be determined. All pixels falling within the region of interest, determined by the rectangle borders, may be put in a new image of (x2 - x1) width and (y2 - y1) height.

## 5.5. Recognition and Classification

The system employs the Artificial Neural Network, specifically the Multilayer Feedforward Network in order to recognize and classify objects. The goal of the Multilayer Feedforward Network is to create a model that correctly maps the input to the output using historical data so that the model can then be used to produce the output when the desired output is unknown.

This type of neural network is known as a supervised network because it requires a desired output in order to learn. Back-Propagation, or the generalized Delta rule, trains the neural network for a given set of input patterns with known classifications. The application of the generalized Delta rule involves two phases, the forward computations and the backward computations. During the first phase, the input is presented and propagated forward through the network to compute the output values for each output unit. The calculated output is compared to the known and desired output, resulting in an error signal for each output unit. The second phase involves a backward pass through the network during which the error signal is passed to each unit in the network, and appropriate weight changes or adjustments are applied.

When each entry of the sample set is presented to the network, the network examines its output response with respect to the input pattern. The output response is then compared to the known and desired output and the error value is calculated. Based on the error, the connection weights are adjusted.

Figure 8 illustrates the necessary steps needed to train the system and calculate the weights.

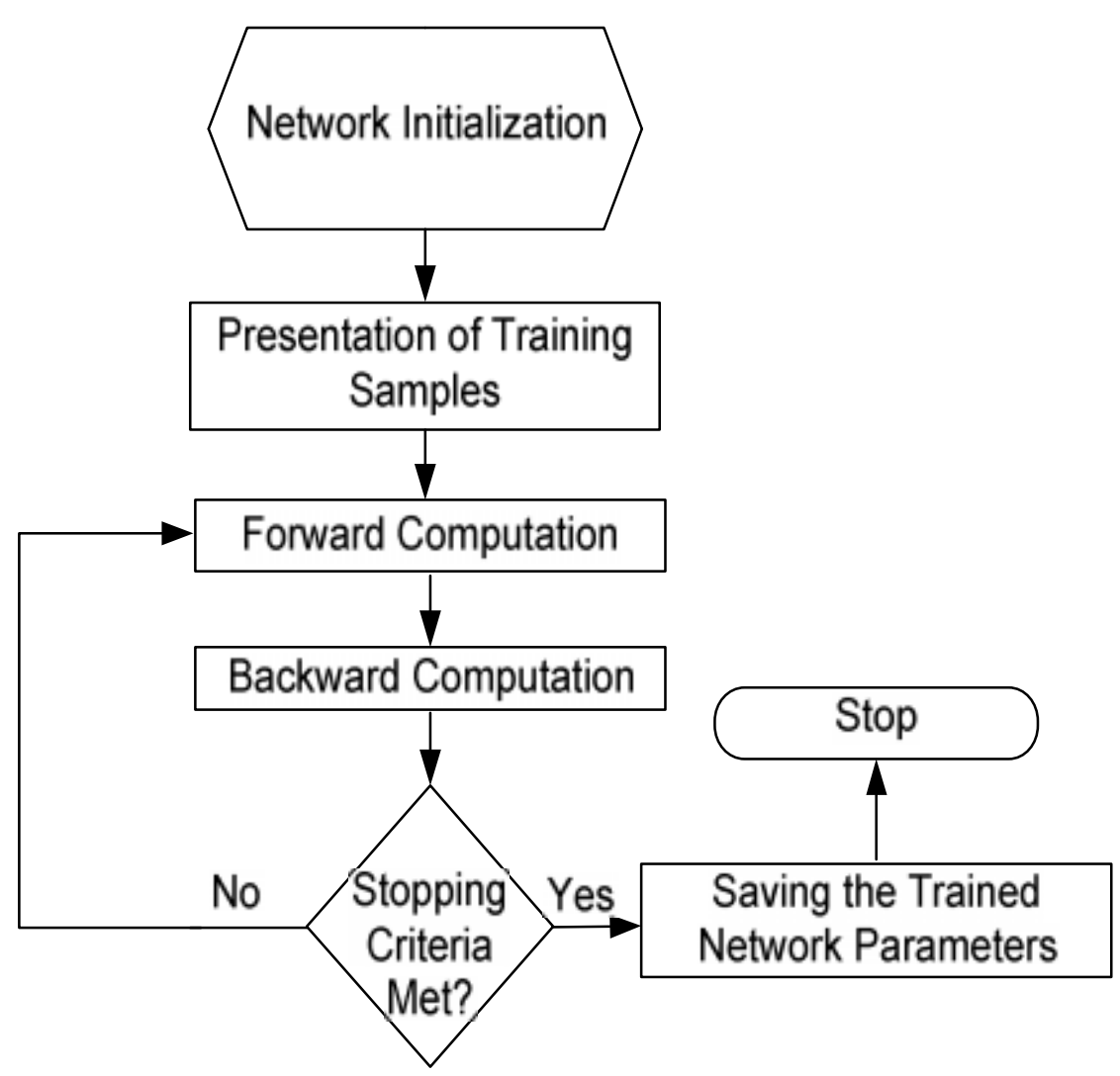

Figure 8. The Back-Propagation algorithm

### 5.5.1. Network Initialization

The neural network has 4,097 inputs, making it fit to accommodate an unvarying image size of 64x64 in addition to one bias value. It is to be noted that a family of colors, such as the yellow and the combinations close to it, cannot be easily identified by a machine using the RGB color space. Moreover, the changes in brightness cause enormous changes in the RGB representation of a certain color.

Therefore, the HSI color space is introduced. HSI color space consists of Hue angle (0 to 360), color Saturation (0 to 100), and Intensity (0 to 255). To be independent of the intensity variance, machine vision applications, such as the landmine detecting robot, will only use the HS space [23].

Colors in the HSI model are defined with respect to the normalized red, green and blue values, given in terms of RGB components by (3), (4), and (5):

$$r = \frac{R}{R+G+B} \tag{3}$$

$$g = \frac{G}{R+G+B} \tag{4}$$

$$b = \frac{B}{R+G+B} \tag{5}$$

Each normalized h, s and i components are then obtained by the following equations:

$$h = \cos^{-1}\left\{\frac{0.5[(r-g)+(r-b)]}{[(r-g)^2+(r-b)(g-b)]^{\frac{1}{2}}}\right\} \quad h \in [0,\Pi]\ for\ b \le g \tag{6}$$

$$h = 2\Pi - \cos^{-1}\left\{\frac{0.5[(r-g)+(r-b)]}{[(r-g)^2+(r-b)(g-b)]^{\frac{1}{2}}}\right\} \quad h \in [\Pi,2\Pi]\ for\ b > g \tag{7}$$

$$s = 1 - 3\min(r,g,b) \quad s \in [0,1] \tag{8}$$

$$i = (R+G+B)/(3\times 255) \tag{9}$$

H, S and I values are then converted respectively into the ranges of [0,360], [0,100], [0, 255], by (10), (11), and (12) respectively.

$$H = h\times 180/\Pi \tag{10}$$

$$S = s\times 100 \tag{11}$$

$$I = i\times 255 \tag{12}$$

Each input, other than the bias, represents an image pixel identified by the aggregate of both the Hue and Saturation components of the pixel; the intensity, as mentioned earlier, is ignored.

There is no general rule that governs the selection of the number of hidden layers in the network or the number of hidden units per layer. Therefore, the system is designed in such a way that the user has the option of defining the number of hidden layers and hidden units in these layers when training the system. The number of output neurons or units is determined by the number of defined classes to be recognized; if three classes are defined, the network will have three output units.

After defining the system's characteristics, the weights are initialized randomly to be between 0.01 and 0.03.

The Multilayer Feedforward Network architecture is designed such that the units in two consecutive layers are fully connected; however, there are no connections between units within one layer. Each hidden unit within the first hidden layer receives input from all input units. Moreover, the hidden units are non-linear units with sigmoid activation functions. The sigmoid activation function is defined in (13).

$$\varphi(v) = \frac{1}{1+\exp(-av)} \tag{13}$$

$v$ represents the aggregate of all input units at the specific hidden unit. Due to the large number of inputs, $v$ is expected to be enormously large. Therefore, $\exp(-av)$ will tend to zero, and consequently $\varphi(v)$ will always be equal to 1. Thus, the input is to be normalized before performing any calculations; moreover, it is necessary to introduce a scale factor that will divide all input values so as to reduce even more their aggregate sum.

Furthermore, in order to reduce the possibility of converging to a local minima rather than converging to the correct global minima, training samples in each iteration are permutated; and training samples of objects belonging to the same class are not provided to the network for training consecutively in the same order for each iteration.

The output nodes are defined to correspond to the number of classes. If three classes are defined, three output nodes will be defined; where class 1 is assigned the output 100, class 2 is assigned the output 010, and class 3 is assigned the output 001.

Forward and Backward calculations are defined in [1].

### 5.5.2. Iteration and Stopping Criteria

The forward and backward computations are repeated as described in the subsections above by presenting new epochs of training examples to the network until the stopping criterion is met. It is to be noted that the stopping criterion is met once the error value is minimized. In general, the Back-Propagation algorithm cannot be shown to converge, and there are no well defined criteria for stopping the network's operation. The Mean Squared Error (MSE), defined in (14), may be considered as stopping criteria and needs to be provided by the user. Unfortunately, this criterion may result in a premature termination of the learning process.

$$MSE = \frac{1}{2}\sum_{j=1}^{N}(Output_j - Desired_j)^2 \tag{14}$$

#### 5.5.3. Running the System for Classification

Once the Neural Network is initialized and trained, the system is ready to be used for classification. In order to run the system for classification, first the image is fed to the processing unit to be filtered and then segmented. Forward computation is carried out on the segmented images and the calculated output classifies the object under investigation.

## 6. RESULTS OF THE SIMULATION

This research aims to provide blueprints for developing an autonomous robot capable of identifying and classifying an object (a landmine, specifically an anti-tank type).

The first step of the classification process is digitally processing the image. Image processing includes image filtering and image segmentation. Figure 9 shows a sample image of two landmines, the TMI-42 and the VS-50. This image was first filtered to eliminate the noise, and then enhanced before being segmented.

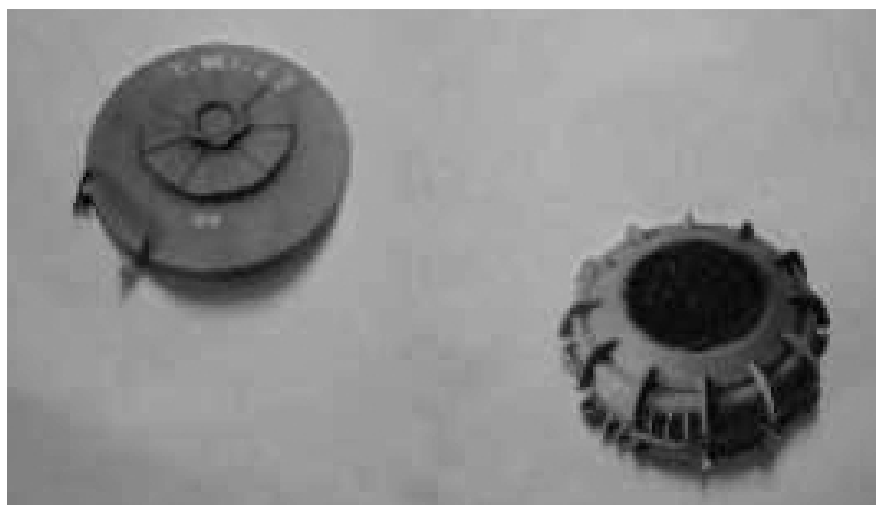

Figure 9. Landmines image to be segmented

For the segmentation process, the user needs to set a threshold value in order for the system to determine the number of classes, or objects, in an image. Assuming a threshold equal to 85, based on the color and pixel location criteria for the segmentation, Figure 10 was segmented into 5 different images, two of which contain a landmine each as shown in Figure 10.

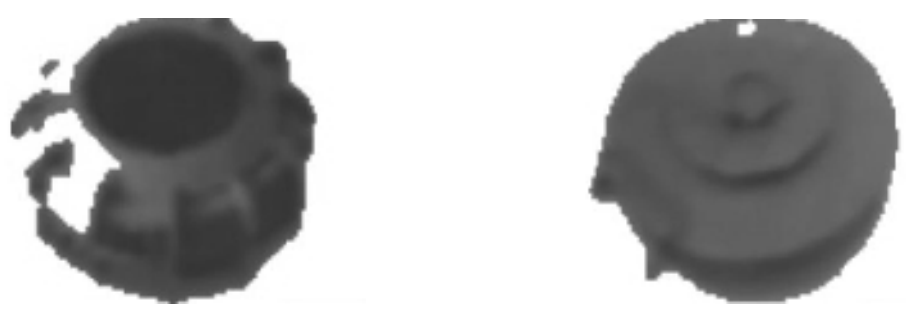

Figure 10. The VS-50 and TMI-42 extracted landmines respectively

It is to be noted that the K-Means clustering is based on iterations and largely depends on the initial guess of the centroids. Two consecutive runs of the K-Means clustering algorithm to segment the same identical image may produce slightly different segmented objects. However, the resulting segmented images would enclose enough information used for the classification when these segmented images are passed through the Neural Network. Moreover, since the K-Means clustering is applied on each pixel of the image for several iterations, running the segmentation algorithm required considerable time to converge and segment the image; therefore, impacting the total time required for the classification process.

Once the images are segmented and the training sets are ready, the implemented system allows the user to select the number of hidden layers, the number of neurons in the hidden layers, the momentum term, the learning rate, and the mean square error for stopping criteria needed to train the system.

Table 3 shows different combinations of the parameters used to train the system as well as some observations regarding the results. These figures were tested for training the system on two classes. It was determined that the best parameters' combination would be to select one hidden layer and 90 neurons per layer, setting the momentum term to 0.2, and the learning rate to 0.01, while the MSE is set to 0.00001. This combination of parameters required an acceptable training time, as compared to other parameters, and resulted in noticeable accuracy rate when running the system.

The system was trained using the above parameters, and was taught to recognize and classify the VS-50 and TMI-42 landmines shown in Figure 10. Once trained, the system was tested on different set of images.

Table 3. Neural network parameters.

| **Hidden Layers** | **Neurons** | **Momentum Term** | **Learning Rate** | **MSE** | **Training Time** | **Accuracy** |
|---|---|---|---|---|---|---|
| 1 | 10 | 0.2 | 0.3 | 0.001 | 1 minutes | Poor |
| 1 | 20 | 0.2 | 0.3 | 0.001 | 1 minutes | Poor |
| 1 | 50 | 0.2 | 0.3 | 0.001 | 2 minutes | Fair |
| 2 | 50 | 0.2 | 0.3 | 0.001 | 3 minutes | Fair |
| 1 | 90 | 0.2 | 0.3 | 0.001 | Diverged | - |
| 1 | 110 | 0.2 | 0.01 | 0.001 | 6 minutes | Good |
| 2 | 90 | 0.2 | 0.01 | 0.001 | 5 minutes | Good |
| 1 | 90 | 0.2 | 0.01 | 0.00001 | 5 minutes | Good |
| 1 | 200 | 0.2 | 0.01 | 0.001 | Diverged | - |
| 5 | 90 | 0.2 | 0.01 | 0.001 | Overnight | - |

Figure 11 represents a graph that illustrates the MSE versus the epoch. As shown, training the system required around 8,000 epochs, or around 5 minutes, in order to optimize the weights and converge to an MSE lower than 0.00001.

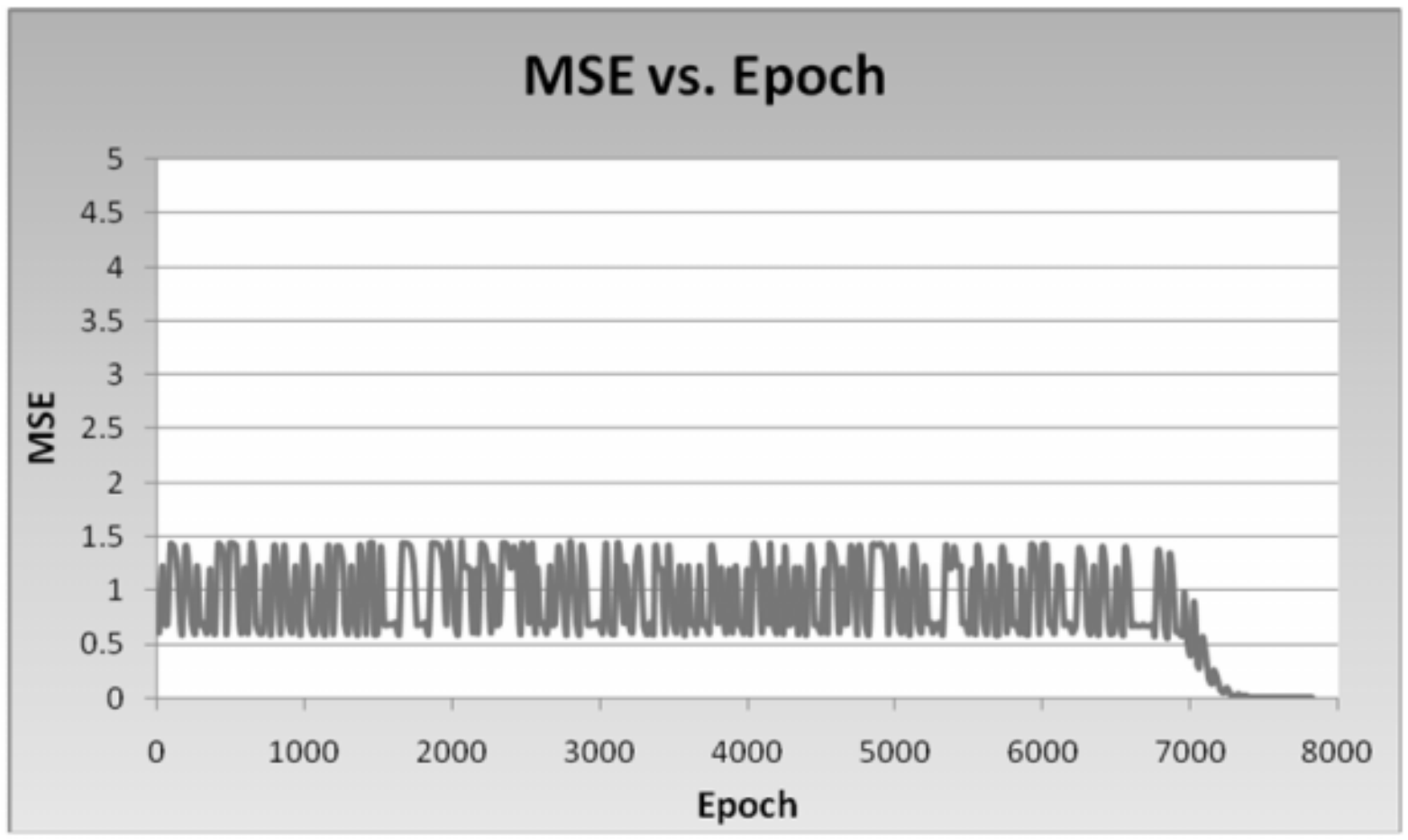

Figure 11. MSE vs. Epoch graph

Figure 12 shows that the trained system is able to identify and classify the VS-50 landmine, marked in a red rectangle, with a correlation factor of 99.6% being the factor determining the proximity of the calculated output of this object to the assigned output of the VS-50 landmine. The system also identified and classified the TMI-42 with a correlation factor of 97.8%. Another image, Figure 13 (a), was also tested. This image is identical to the image of Figure 12; however, the landmines are placed differently. The trained system was able to identify and classify the VS-50 landmine, marked in a red rectangle, with a correlation factor of 98.6%. The system also identified and classified the TMI-42 with a correlation factor of 93.3%.

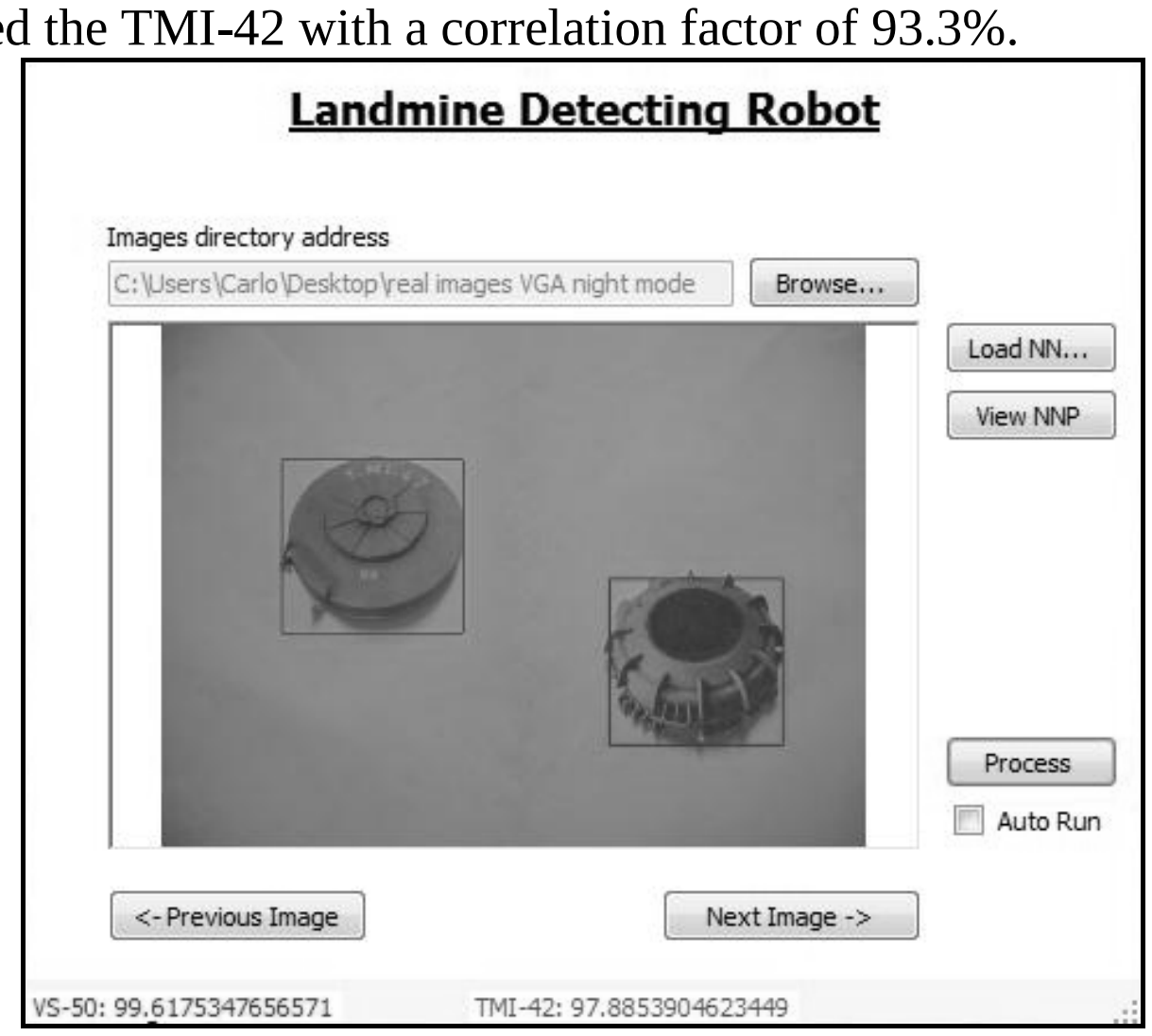

Figure 12. The VS-50 and TMI-42 landmines classification

It is worth noting that both runs on Figure 12 and Figure 13 (a), the K-Means clustering technique extracted correctly and precisely the exact landmines shapes. However, Figure 13 (b) shows a distorted identification and classification of the TMI-42 landmine; whereby the system had identified and classified the TMI-42 landmine with a low correlation factor of 69.5%. These results are due to a distorted segmentation, for the exact shape of the TMI-42 was not extracted and passed on to the network for classification; however, the system was still able to classify this object as the TMI-42 landmine.

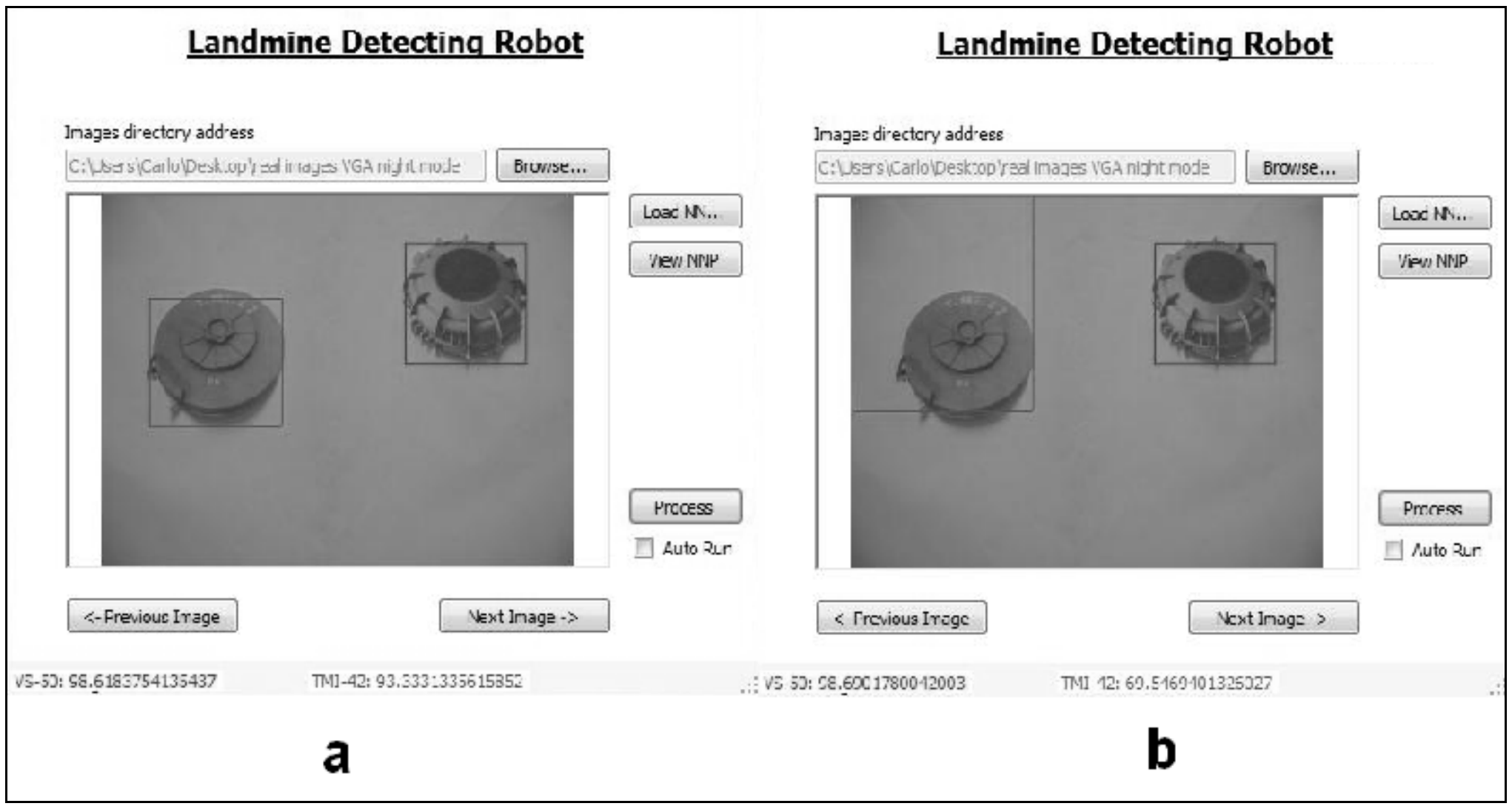

Figure 13. (a) Displaced VS-50 and TMI-42 landmines classification. (b) Distorted identification and classification of the TMI-42

In addition, two scenarios were tested: Figure 14 (a) simulates the VS-50 landmine being rotated, and Figure 14 (b) simulates a part of the VS-50 being covered by a sheet of paper. Running Figure 14 (a), the system was able to identify and classify the TMI-42 landmine, marked with a red rectangle, with a correlation factor of 89.4%.

Although the system was not trained to detect a rotated version of the VS-50 landmine, it was able to identify and classify the VS-50, but, with a lower correlation factor of 73.3%. This correlation factor may be increased if a rotated version of the VS-50 landmine is also added to the training set.

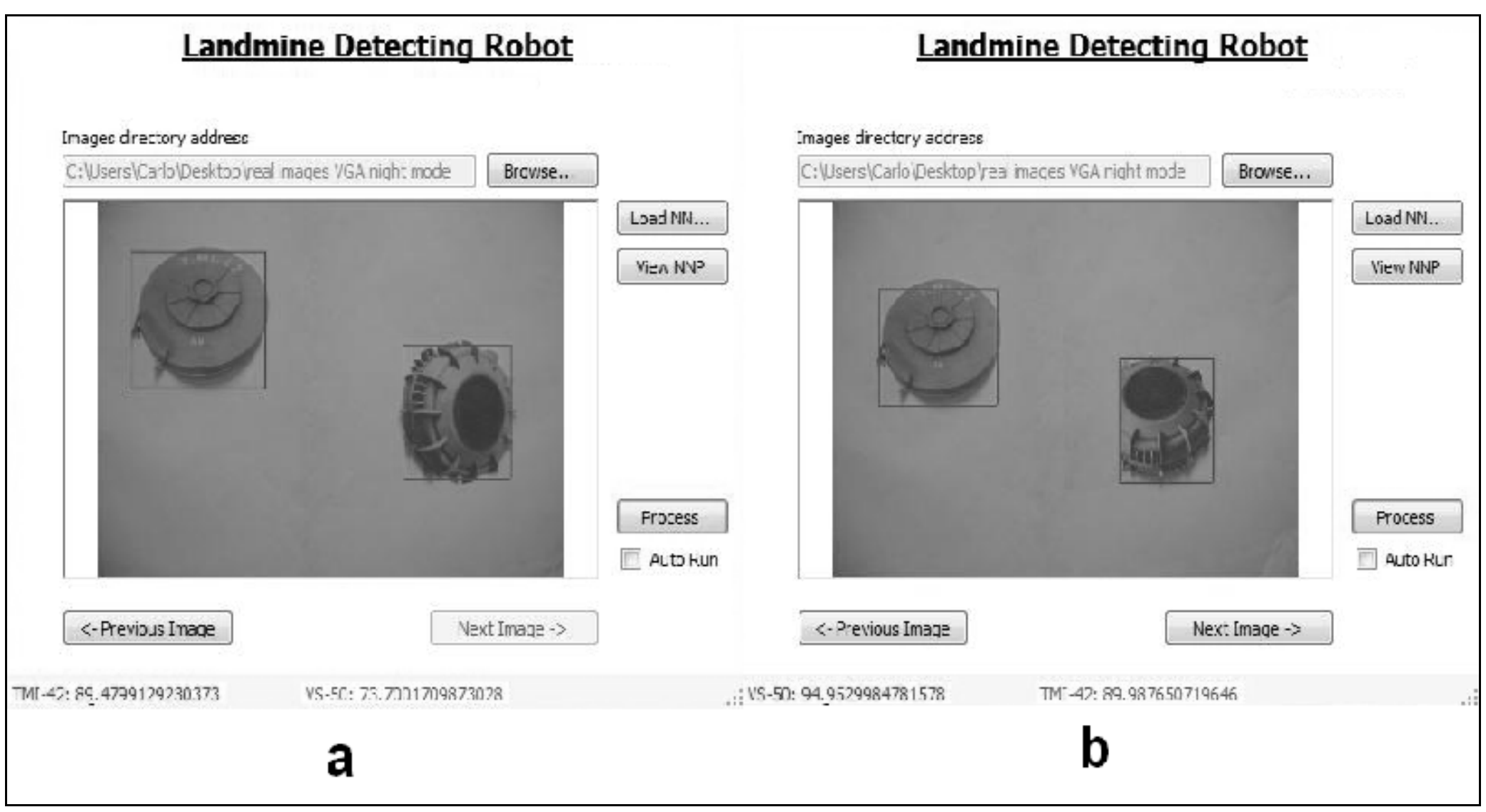

Figure 14. (a) Rotated VS-50. (b) Covered VS-50

Running Figure 14 (b), the system was able to identify and classify the TMI-42 landmine, marked with a red rectangle, with a correlation factor of 89.9%. The system was also able to identify and classify the VS-50 landmine, covered with a sheet, with a correlation factor of 94.4% compared to 99.6% correlation factor for the completed version of the VS-50.

## 7. Conclusion and Perspective

This research's aim is to provide blueprints for developing a system, an autonomous robot, capable of identifying and classifying an object (a landmine specifically an anti-tank, a cluster bomb or unexploded ordnance).

This paper investigated different sensing techniques, image processing techniques, and recognition and classification techniques.

Based on the constraints and specifications required for the system, the best techniques were adopted for the design and implementation of the robot system: the designed system employs a camera, as a sensing technique, to capture real-time images of the scanned area. The captured images are then fed into a computing unit to be digitally processed. The processed images are then input into a Neural Network to be classified. The developed system is implemented using C# under the Microsoft .Net Framework and a laptop. Several complications were encountered during the course of the implementation of the system, specifically:

- Due to the large number of input, the output at the neuron was always calculated to be 1. Therefore it was necessary to normalize the input before performing any calculations; moreover, it was necessary to introduce a scale factor that divides all input values to reduce their aggregate even more.
- The network was not well trained, and the stopping criteria was not met for adopting a single output that points to the class of the object under study. However, it was determined that defining outputs nodes corresponding to the number of classes allow the system to converge.

- The captured images are represented in the RGB color space. However, the changes in brightness cause enormous changes in the RGB representation of a certain color and thus will cause the machine (robot) to inaccurately classify an object. Hence, the HSI color space, which consists of Hue angle, color Saturation and intensity, is introduced. To be independent of the intensity variance, machine vision applications only use the HS space.

Once the complications were overcome, the system proved to be able to identify and classify different types of landmines under various conditions, such as a rotating landmine or a partially covered one. Table 4 shows the different examples simulated in this study, and lists the average correlation factors that the system determined for the classified objects

This research aims to provide blueprints for developing an autonomous robot capable of identifying and classifying an object (a landmine, specifically an anti-tank type).

Table 4. Simulation summary.

| | **Landmine Scenario** | | | |
|---|---|---|---|---|
| | **TMI-42** | **VS-50** | **Covered VS-50** | **Rotated VS-50** |
| **Average Correlation Factor in Percentage** | 86.04% | 98.97% | 90.00% | 70.00% |

Study results have proven that the system has the ability to classify the object with around 90% accuracy rate.

In few instances, the system was not able to identify correctly the landmine; this drawback was mainly due to the segmentation process where the K-Means clustering algorithm failed to effectively extract the exact objects' platforms from the image.

Per this research, a camera is used to capture colored images of the area under investigation. These captured images consist of a two dimensional representation of the surface. Two dimensional images provide plenty of data to analyze, but it does not release information about the depth, the orientation and the placement of an object; these additional records may provide more accuracy on the classification process. The latter information may be retrieved by using two cameras placed next to each other to capture images of the same region. These captured images are used to create a disparity image representing a three dimensional view of the surface.

Moreover, and in order to reduce the overhead processing, the robot needs a triggering device to notify the processing unit on when to capture and process an image and therefore identify and classify the landmines if any. Since landmines contain metal materials, a metal detector may be employed. The autonomous robot is then programmed to sweep the ground under investigation. Once the metal detector senses a suspicious object, the processing unit is notified and in turn triggers the robot to stop and the camera to capture an image of the scanned area. Afterwards, the image is processed per the methodology described throughout this study.

Furthermore, and as previously stated, this research focused on classifying anti-tank landmines that are laid on the surface of earth; the metal detector may also be used to detect landmines buried under the earth; however, the camera will be useless under this scenario where an ultra sonic sensor may be used in addition to the metal detector in order to provide information adequate to classify the object.

## REFERENCES

[1] R. Achkar, M. Owayjan, and C. Mrad "Landmine Detection and Classification Using MLP", IEEE Third International Conference on Computational Intelligence Modeling and Simulation, CIMSim 2011, Langkawi Malaysia, September 20- 22, pp 1-6.

[2] New Internationalist Magazine. (1997-1998). “Landmines-The Facts”. Available: http://www.newint.org/easier-english/landmine/lmfacts.html

[3] United Nations Mine Clearance and Policy Unit, Department of Humanitarian Affairs. (1997, September). “Landmines Factsheet”. Available: http://www.un.org/cyberschoolbus/banmines/facts.asp

[4] The United Nations – Department of Humanitarian Affairs. (1996, August). Available: http://www.un.org/Depts/dha/

[5] “South Lebanon Demining Initiative”, Statement by UNIFIL Force Commander, 4 February 2002

[6] International Campaign to Ban Landmines. (2003). “Landmine Monitor Report 2003: Toward a Mine-Free World”, Lebanon. Available: www.icbl.org/lm

[7] United Nations Association of the United States of America. "Landmines Overview". Available: http://www.landmines.org/Page.aspx?pid=789

[8] Canadian Red Cross. (2008, August). “The Landmine Epidemic”. Available: http://www.redcross.ca/article.asp?id=1945&tid=006

[9] Wikipedia. (2010, June). “Demining”. Available: http://en.wikipedia.org/wiki/Demining

[10] Geneva International Centre for Humanitarian Demining. (1999). “Technology”. Available: http://www.gichd.org/links-information-database/technologies/linksdb/technologies/manual-demining/

[11] Geneva International Centre for Humanitarian Demining , A Study of Mechanical Application in Demining, Geneva: May 2004

[12] C. Bruschini, B. Gros, "A Survey on Sensor Technology for Landmine Detection", Journal of Humanitarian Demining, Issue 2.1, February 1998

[13] J. Paik, C.P. Lee, M.A. Abidi, "Image Processing-Based Mine Detecting Techniques: A Review", Received October 2, 2011, revised February 8, 2002, pp. 156-169

[14] A. K. Jain, Fundamentals of digital image processing, United States of America: Prentice Hall, 1989.

[15] R.C. Gonzales and R.E. Woods, Digital Image Processing, 2nd ed, United States of America: Prentice Hall, 2001.

[16] A.T. Sapkal, C. Bokhare and N.Z. Tarapore, "Satellite Image Classification using the Back Propagation Algorithm of Artificial Neural Network", Technical Article, pp. 1-4

[17] S. Haykin, Neural Networks - A Comprehensive Foundation, 2nd ed., India: Prentice Hall, 1998.

[18] S. Russel and P. Norvig, Artificial Intelligence A Modern Approach, 3rd ed., United States of America: Prentice Hall, 2009.

[19] B. Jahne, Digital Image Processing, 6th revised and extended edition, Netherland: Springer, 2005

[20] P. Sudhakara Rao, A. Gopal, S. Md. Iqbal, R. Revathy and K. Meenakshi, "Classification of Fruits Based on Shape Using Image-Processing Techniques", n.d.

[21] T. Kanungo, D.M. Mount, N.S. Netanyahu, C.D. Piatko, R. Silverman, and A.Y. Wu, "An Efficient k-Means Clustering Algorithm: Analysis and Implementation", paper in IEEE Transactions on Pattern Analysis and Machine Intelligence, Issue 7, July 2002

[22] S.J. Redmond, "A method for initialising the K-means clustering algorithm using kd-trees", 2005, pp. 1-4

[23] P. Sudhakara Rao, A. Gopal, S. Md. Iqbal, R. Revathy and K. Meenakshi, "Colour Analysis of Fruits Using Machine Vision System for Automatic Sorting and Grading", n.d.

## Authors

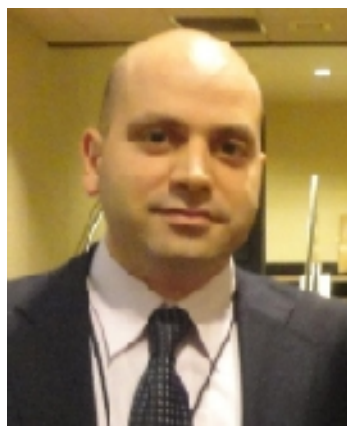

**Roger Achkar** received the B.E. degree in Electrical Engineering from the Lebanese University, faculty of engineering II in 2002. He obtained the M.S. degree in Industrial Control from the Lebanese University in 2004 and a Ph.D. degree in Automation and Control from the UTC University- Compiegne- France in 2008. In the same year, he joined the American University of Science & Technology-Lebanon as Assistant Professor and coordinator of the CCE and CSI departments in Sidon and Zahle Campuses. Today, Dr. Achkar is the General Academic Coordinator for AUST- Zahle Campus. Dr Achkar is a Member of the IEEE Engineering in the Computational Intelligence Society and a Member of the Order of Engineers and Architects-Lebanon. His Research interests are in Electrical and Mechanical Engineering Education. He has been involved in the implementation of artificial neural networks (multi-layer-perceptrons, radial basis function, support vector machine, torque method, adaptive Filtering using Kernel Methods) on different systems such as: robot mine detector, active magnetic bearing AMB and artificial vision systems.

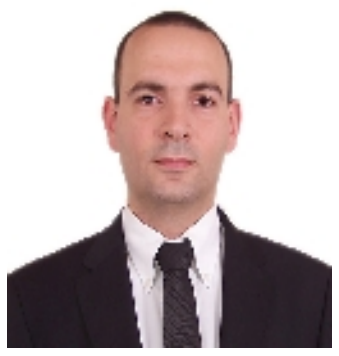

**Michel Owayjan** (IEEEM'06) received the B.E. degree in electrical engineering and the M.E. degree in engineering management from the American University of Beirut, Lebanon, in 1999 and 2002 respectively. Since 1999, he has been working as an electrical engineer consultant for Ouayjan Architecture, Lebanon. During 2000-2004, he taught as a part-time instructor at the American University of Science & Technology (AUST), Beirut, Lebanon. Since 2005, he is serving as a lecturer and a coordinator in the departments of Computer and Communications Engineering, Computer Science, and Biomedical Engineering at AUST, Beirut, Lebanon. Mr. Owayjan is a member in the IEEE Computer Society since 2006. During 2009-2010 he served as the membership development and industry relations officer of the Computer Society in the IEEE Lebanon section. Since 2011 he is serving as the Secretary of the Computer Society in the IEEE Lebanon section. He is the counselor of the AUST IEEE Student Branch since 2010. His research interests include signal processing, system modeling and simulation, optimization, operations research, control, algorithms, neural networks, and artificial intelligence.